\documentclass[journal]{IEEEtran}
\usepackage[T1]{fontenc}   
\usepackage{makecell}
\usepackage{color}
\usepackage{booktabs}
\usepackage{graphicx}
\usepackage{hyperref}  
\hypersetup{hidelinks,colorlinks=true,linkcolor=red,citecolor=blue}
\usepackage{cite}
\usepackage{comment}
\usepackage{amsfonts,amssymb}  
\usepackage{colortbl}
\usepackage{stfloats}
\usepackage{bm}  
\usepackage{amsmath}
\usepackage{threeparttable}
\usepackage{soul}
\soulregister\cite7
\soulregister\ref7
\usepackage{multirow}

\usepackage{array}
\usepackage{booktabs}
\usepackage{makecell}
\usepackage{multirow}
\usepackage{graphicx}
\usepackage[justification=centering]{caption}
\usepackage{amsmath}
\usepackage{algorithm} 
\usepackage{algpseudocode} 

\usepackage{pgfplots}
\pgfplotsset{compat=1.17} 
\usepackage{ulem}

\usepackage{tabularx} 
\usepackage{ulem} 

\begin{document}

\title{\textbf{MambaSOD: Dual Mamba-Driven Cross-Modal Fusion Network for RGB-D Salient Object Detection}}
\author{Yue Zhan, Zhihong Zeng, Haijun Liu, Xiaoheng Tan, Yinli Tian
\thanks{This paper was supported by the National Natural Science Foundation of China under Grant U20A20157. Corresponding author: Zhihong Zeng.}
\thanks{Y. Zhan is with the Department of Electrical and Electronic Engineering, the University of Hong Kong, Hong Kong SAR, China.}
\thanks{Z. Zeng, H. Liu and X. Tan are with the School of Microelectronics and Communication Engineering, Chongqing University, Chongqing 400044, China (e-mail: azhihong@cqu.edu.cn).} 
\thanks{Yinli Tian is with the School of Software Engineering, Chongqing University of Posts and Telecommunications, Chongqing 400065, China.}
}
\date{}   

\markboth{}%
{Shell \MakeLowercase{\textit{et al.}}: A Sample Article Using IEEEtran.cls for IEEE Journals}

\maketitle

\begin{abstract}
The purpose of RGB-D Salient Object Detection (SOD) is to pinpoint the most visually conspicuous areas within images accurately. While conventional deep models heavily rely on CNN extractors and overlook the long-range contextual dependencies, subsequent transformer-based models have addressed the issue to some extent but introduce high computational complexity. Moreover, incorporating spatial information from depth maps has been proven effective for this task. A primary challenge of this issue is how to fuse the complementary information from RGB and depth effectively. In this paper, we propose a dual Mamba-driven cross-modal fusion network for RGB-D SOD, named MambaSOD. Specifically, we first employ a dual Mamba-driven feature extractor for both RGB and depth to model the long-range dependencies in multiple modality inputs with linear complexity. Then, we design a cross-modal fusion Mamba for the captured multi-modal features to fully utilize the complementary information between the RGB and depth features. To the best of our knowledge, this work is the first attempt to explore the potential of the Mamba in the RGB-D SOD task, offering a novel perspective. Numerous experiments conducted on six prevailing datasets demonstrate our method's superiority over sixteen state-of-the-art RGB-D SOD models. The source code will be released at \href{https://github.com/YueZhan721/MambaSOD}{https://github.com/YueZhan721/MambaSOD} 

\end{abstract}

\begin{IEEEkeywords}
RGB-D Salient Object Detection, State Space Model, Mamba-based Backbone, Cross-modal Fusion.
\end{IEEEkeywords}

\section{INTRODUCTION}
\IEEEPARstart{S}{alient} object detection (SOD) constitutes a task in computer vision, focusing on the identification of the most prominent objects in an image or video scene. This task has significant implications in the image and video processing fields, such as video analysis \cite{fan2019shifting}, visual tracing \cite{ ZHANG2020107130}, and image quality evaluation \cite{yang2021reference}.

Although significant advances have been made in the RGB SOD field \cite{8793227,zhao2019egnetedge,8953756}, detection performance notably decreases when objects are blended with similar-looking surroundings or in cluttered backgrounds. To address this challenge, depth maps are utilized as supplementary input to enhance the comprehension of the images' spatial cues. In this way, some research \cite{8638984,9466383,9076277} shows that incorporating depth information as an additional resource achieves exceptional RGB-D SOD performance in demanding scenes. Therefore, the valid feature fusion of RGB and depth is critical for salient object detection. 

\begin{figure}[t]
	\centering
        \captionsetup{
        justification=justified,
        singlelinecheck=false 
    }
        \includegraphics[width=0.98\linewidth]{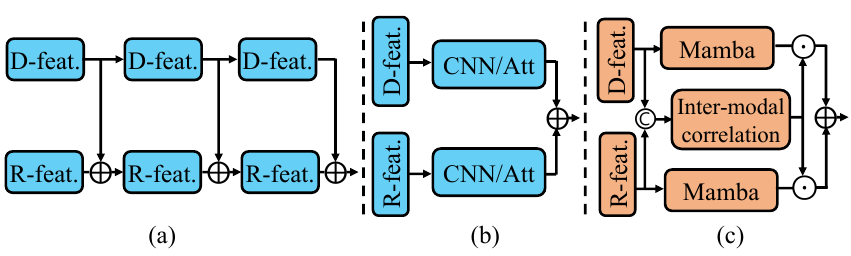}  \\
	\caption{Cross-modal feature fusion strategies of existing RGB-D SOD methods ((a)\cite{zeng2023compensated,fan2020bbs,wu2021mobilesal,2023DepthInjection,9686679} and (b)\cite{zhang2021bts,2023HiDAnet,zhong2024magnet}) as well as the proposed cross-modal fusion Mamba (c). `D-feat.' and `R-feat.' represent depth and RGB features, respectively.}
	\label{fig:feature_fusion_strategy}
\end{figure}

So far, many RGB-D feature fusion methods for salient detection models have been proposed\cite{fan2020bbs,wu2021mobilesal,2023DepthInjection,9686679,zeng2023compensated,10184101,zhang2021bts,2023HiDAnet,zhong2024magnet}. For example, DIF\cite{2023DepthInjection} proposed to directly inject the depth cues into the RGB features, while HiDAnet \cite{2023HiDAnet} adopted spatial attention and channel attention to enhance self-modal features for subsequent fusion stage. Overall, the current RGB-D feature fusion strategies mainly have two paradigms. The first one is injecting the depth features to its counterpart (as shown in Fig. \ref{fig:feature_fusion_strategy} (a)) while the second one is adopting convolution neural network, channel, or spatial attention to enhance self-modal features for subsequent RGB-D feature fusion stage (as depicted in Fig. \ref{fig:feature_fusion_strategy} (b)). 
However, the works mentioned above\cite{fan2020bbs,wu2021mobilesal,2023DepthInjection,9686679,zeng2023compensated,zhang2021bts,2023HiDAnet,zhong2024magnet} focus less on researching the long-range dependencies modeling during the cross-modal feature fusion. Therefore, we propose a Cross-Modal Fusion Mamba (CMM) which is effective for inter-modal correlation's long-range dependencies modeling and modality-specific feature enhancement during the RGB-D feature fusion, as depicted in Fig. \ref{fig:feature_fusion_strategy} (c).

Furthermore, the convolutional neural network (CNN) is very popular as a critical backbone to extract multi-level features in the RGB-D SOD task. For example, VGG \cite{simonyan2014very} and ResNet \cite{he2016deep} allow the model to capture refined and rich local feature details. Newer architectures like DenseNet \cite{huang2017densely} and EfficientNet \cite{tan2019efficientnet} have further enhanced feature extraction capability and network performance. MobileNet \cite{MobileNet} makes it possible to deploy the CNN on edge devices. Despite these advances, the convolutional operations are restricted to local feature extraction and have limited ability to model global information.

In that case, Transformer-based backbone \cite{dosovitskiy2020image, liu2021swin,liu2021tritransnet} has been introduced to the RGB-D SOD field for long-range dependencies modeling. For example, inspired by the work of Vision Transformer \cite{dosovitskiy2020image} and Swin Transformer \cite{liu2021swin}, SwinNet \cite{9611276} and MITF-Net\cite{9925217} adopted transformer blocks in the backbone to extract more global features. GroupTransNet\cite{fang2024grouptransnet} proposed a group transformer for long-range dependencies learning of cross-layer features. However, models based on Transformers or a variant of Transformers suffer from significant computational complexity due to the quadratic growth of resources with the number of tokens stemming from the self-attention mechanism.

Therefore, we propose a dual Mamba-driven cross-modal fusion network for RGB-D SOD, named MambaSOD. Unlike the previous methods that adopted these feature extraction networks based on CNN or Transformer, we are the first to introduce a dual Mamba backbone in the RGB-D SOD field. The main reason is that the Mamba-based backbone\cite{liu2024vmamba} has the ability to model long-range dependencies within images while maintaining linear complexity. More importantly, to address the issue of effective cross-modal features fusion, we proposed a cross-modal fusion Mamba to model inter-modal correlation's long-range dependencies and enhance modality-specific features, so that the two modalities have sufficient interaction to learn the complementary features for RGB-D feature fusion. Finally, we adopt a multi-level refinement module as our decoder to predict the accurate salient maps.

Our key contributions are outlined below:
\begin{itemize}
    \item We propose a dual Mamba-driven cross-modal fusion network for RGB-D SOD, MambaSOD, which utilizes a dual Mamba-based backbone for the feature extraction of RGB and depth. As far as we know, we are the first to introduce the Mamba-base backbone into the RGB-D SOD field and numerous experiments have proven its effectiveness.
    \item We propose a cross-modal fusion Mamba module that can enhance modality-specific features and inter-modal correlation's long-range dependencies modeling. Specifically, in inter-modal correlation's long-range dependencies modeling, we project features from two modalities into a shared space to achieve complementary feature learning, which is crucial to RGB-D feature fusion.
    \item We perform extensive experiments across multiple popular datasets, and the results show our proposed MambaSOD achieves excellent performance, which proves the validity and superiority of our method.
\end{itemize}

\section{RELATED WORK} \label{Section_2}
\subsection{Backbone of RGB-D SOD}
In the field of RGB-D SOD, it is crucial to effectively extract the features of different modalities. Initially, the feature extraction backbone was primarily a single-stream network. Fan et al. \cite{fan2020rethinking} used a single-stream backbone to extract and fuse features, combined with a doorway mechanism to tackle the issue of poor quality of some depth maps. Fu et al. \cite{JL-DCF} chose a shared CNN backbone to obtain related features of RGB and depth, aiming to realize the purpose of cross-modal information-sharing via joint learning. However, while single-stream networks are easier to train due to lower network complexity and fewer parameters, they may also encounter challenges in extracting cross-modal features. For example, when the RGB and depth information are highly varied, it may be difficult for a single network to optimally process these different types of data simultaneously.

To tackle the issues of single-stream backbone networks, some researchers have proposed dual-stream backbone networks. To sum up, the methods of dual-stream backbone networks are divided into CNN-based methods and Transformer-based ones. In the initial network, Li et al. \cite{7780427} simply stacked convolutional kernels and maximal pooling layers following the related work of VGG \cite{VGG}, which enables the network to efficiently capture more complex features while increasing the depth layer by layer. At almost the same time, He et al. \cite{ResNet} proposed the ResNet network, in which the concept of residual learning was first introduced. By using skip connections, information is allowed to be passed directly from shallower to deeper layers, and this can solve the degradation problem during deep network training. Currently, many RGB-D SOD models utilize ResNet as the backbone. For example, BBSNet \cite{9405467} conducted a dual-stream backbone based on ResNet50 to extract RGB and depth features, respectively. Later, considering mobile and edge devices, inspired by the work of MobileNet \cite{MobileNet}, Wu et al. \cite{wu2021mobilesal} introduced MobileNetV2 into RGB-D SOD as feature extraction network, which reduces computation and model size while maintaining reasonable performance. Nevertheless, these methods are all modeled with CNN layers and the lack of global modeling capability leads to performance limitations on SOD tasks. 

The successful example of transformers in Natural Language Processing has motivated researchers to assess its feasibility in image processing. Vision Transformer \cite{dosovitskiy2020image} splits the image into multiple fixed-size patches and then captures the complex relationships between the patches through a self-attention mechanism. In this way, any region can be dynamically attended to over the entire image range, thus effectively capturing global dependencies. Swin Transformer \cite{liu2021swin} aggregates information at different levels by restricting self-attention to computations within localized windows and then gradually expanding the size of these windows, which aims to reduce the computation while retaining the Transformer's advantage of modeling global information. Pyramid Vision Transformer \cite{PVT} designed a progressive shrinking pyramid and spatial-reduction attention to enable high-resolution dense prediction while effectively reducing the model computation to some extent. Nevertheless, the attention mechanism results in Transformer-based methods having a huge number of parameters and quadratic computational complexity.

To this end, inspired by the Mamba architecture, known for its outperformance in modeling local and global contextual information while keeping computational efficiency as a State Space Model (SSM), we introduce a backbone network based on Mamba.
\begin{figure*}
    \captionsetup{
    justification=justified,
    singlelinecheck=false 
    }
    \centering
    \includegraphics[width=1\linewidth]{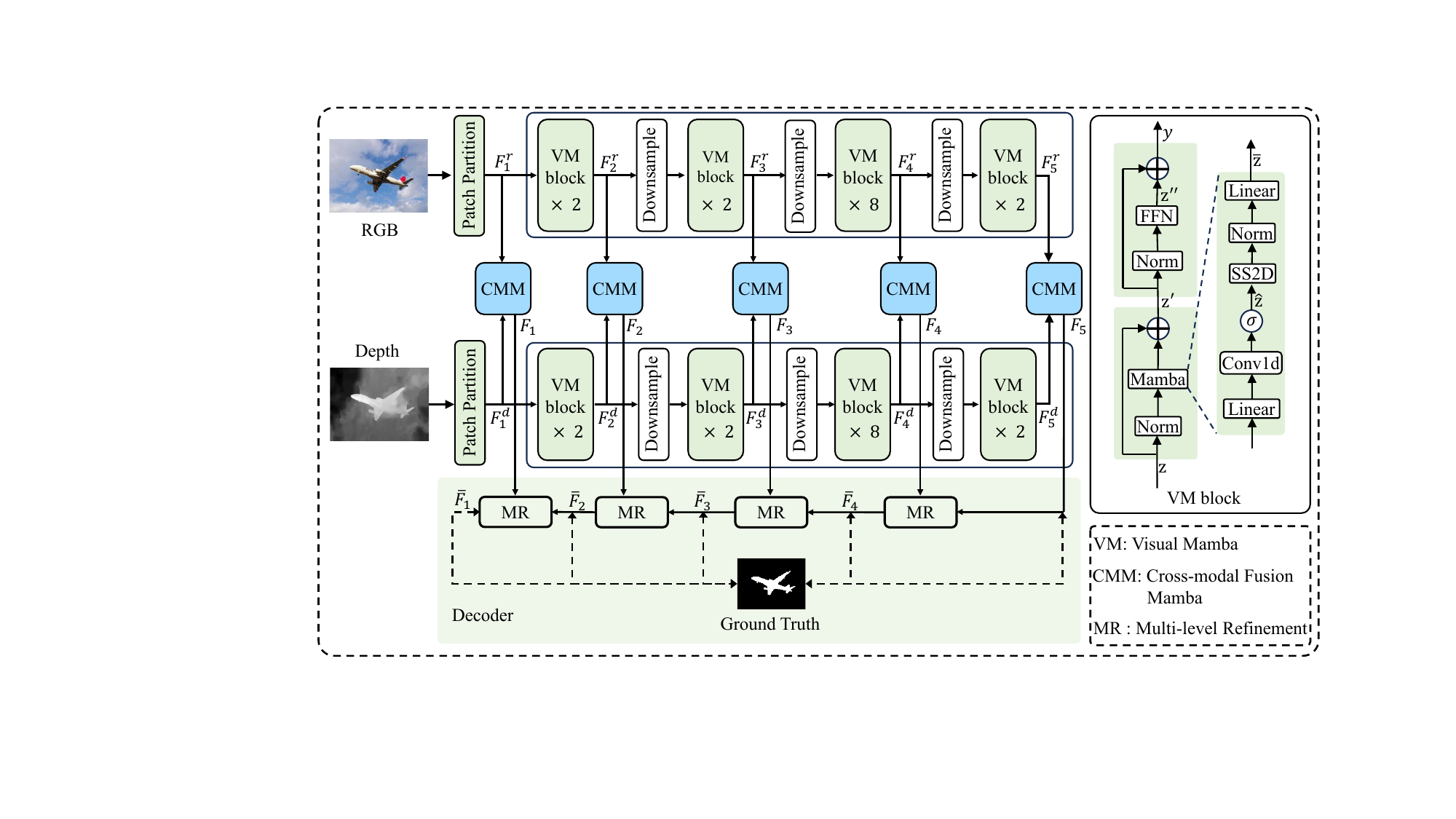}  \\
    \caption{The pipeline of our MambaSOD. It includes the Mamba-based encoder, the cross-modal fusion Mamba, and the multi-level refinement decoder.}
    \label{fig:pipeline}
\end{figure*}
\subsection{Cross-modal feature fusion in RGB-D SOD}
Effective cross-modal feature fusion is particularly important to improve the performance in RGB-D SOD. Existing cross-modal feature fusion models are typically classified into three kinds: early fusion, middle fusion, and late fusion. The early fusion merges the inputs of the two modalities by concatenation or other methods before the feature extraction network. For example, in JL-DCF \cite{JL-DCF}, the initial RGB and depth maps are first concatenated together, and the subsequent feature extraction backbone simultaneously obtains cues from the combined RGB-D. Late fusion refers to extracting features of different modalities separately, then fusing the two branches of features, and finally predicting the saliency map. Peng et al. \cite{peng2014rgbd} performed separate saliency predictions by RGB and depth, and then fused the two predictions for the final saliency map. The middle fusion stage can be further divided into two types, as shown in Fig. \ref{fig:feature_fusion_strategy}, the (a) is to simply concatenate the RGB and depth features, and the (b) is to enhance the self-modality feature with CNN or self-attention before performing RGB-D fusion. For instance, \cite{9405467} straightforwardly fuse the RGB and depth features by element-wise addition to achieve the goal of feature fusion. Wang et al. \cite{wang2022learning} first calculated the respective correlation maps of RGB and depth, and then conduct them by pixel-wise addition to interact cross-modal features. A different example is that \cite{BTS} conducted self-enhancement on extracted features through the channel attention mechanism before feature fusion. 

However, the above middle fusion approaches mainly adopt CNN or attention mechanisms for feature enhancement and fusion, lacking the exploration of inter-modal correlation's long-range dependencies. Hence, we propose a cross-modal fusion Mamba, as depicted in Fig. \ref{fig:feature_fusion_strategy}(c), which aims to achieve the self-modal enhancement and also learn the inter-modal correlation as feature complementarity.

\section{METHODOLOGY} \label{Section_3} 
\subsection{Preliminaries}
\textbf{Formulation of SSM.} Originating from the Kalman filter \cite{kalman1960new}, the State Space Model (SSM) can be considered as a linear system that transfers the input stimulation $ x(t) \in \mathbb{R}$ to output response $ y(t) \in \mathbb{R}$ through the hidden state $ h(t) \in \mathbb{R}^{N}$. Concretely, continuous-time SSM can be expressed as a linear Ordinary Differential Equation (ODE), which is calculated as,
\begin{equation}
\begin{aligned}
    h(t) &= \mathbf{A} h(t-1) + \mathbf{B} x(t),\\
    y(t) &= \mathbf{C} h(t),
\end{aligned}
\label{eq1}
\end{equation}
in which the weighting parameters $ \bm{A} \in \mathbb{R}^{N \times N}$, $ \bm{B} \in \mathbb{R}^{N \times 1}$, and $ \bm{C} \in \mathbb{R}^{1 \times N}$.

\textbf{Discretization of SSM.} To be integrated into deep models, continuous-time SSM must undergo discretization in advance, like Structured State Space Sequence Models (S4) \cite{gu2021efficiently} and Mamba \cite{gu2023mamba}. Concretely, a timescale parameter $\mathbf{\Delta}$ is used to transfer the continuous $\mathbf{A}, \mathbf{B}$ to discrete version $\mathbf{\bar{A}}, \mathbf{\bar{B}}$. A widely conducted method is Zero-Order Hold (ZOH), which can be defined below:
\begin{equation}
\begin{aligned}
    \bar{\mathbf{A}} &= \exp(\mathbf{\Delta A}),
    \\
    \bar{\mathbf{B}} &= (\mathbf{\Delta A})^{-1} (\exp(\mathbf{\Delta A}) - \mathbf{I}) \cdot \mathbf{\Delta B}.
\end{aligned}
\label{eq2}
\end{equation}
After discretization, the whole SSM system can be reshown as:
\begin{equation}
\begin{aligned}
    h_t &= \bar{\mathbf{A}} h_{t-1} + \bar{\mathbf{B}} x_t,
    \\
    y_t &= \mathbf{C} h_t.
\end{aligned}
\label{eq3}
\end{equation}
Furthermore, the convolution representation of the final discreet SMM model is defined as:
\begin{equation}
\begin{aligned}
    \mathbf{\bar{K}} &= (\mathbf{C\bar{B}}, \mathbf{C\bar{A}\bar{B}}, \ldots, \mathbf{C\bar{A}^{L-1}\bar{B}}),
    \\
    \mathbf{y} &= \mathbf{x} * \mathbf{\bar{K}},
\end{aligned}
\label{eq4}
\end{equation}
where the input $\mathbf{x}$ has $\mathbf{L}$ tokens, and $\mathbf{\bar{K}} \in \mathbb{R}^{L}$ represents a $1D$ convolutional kernel. To facilitate the expression, given an input $\mathbf{x}$, the whole SSM operation can be simply defined as:
\begin{equation}
\begin{aligned}
    \mathbf{y} &= \text{SSM}(\mathbf{x})
\end{aligned}
\end{equation}

\textbf{2D-Selective-Scan (SS2D).}
Inspired by the success of the S6 scanning operation \cite{gu2023mamba} in natural language processing, the selective scan approach is adopted into image processing. However, due to 2D images being non-sequential and possessing spatial information, an SS2D module \cite{liu2024vmamba} was proposed to adapt S6 for vision data without compromising its advantages, which enables the obtaining of not only local but also global features. Specifically, a 2D image input is first divided into several patches and then scanned in four directions to produce four sequences that are processed by SSM individually. Finally, a merging operation is performed on the processed sequences to reconstruct the patches into a 2D image again. The SS2D process is defined as: 
\begin{equation}
\begin{aligned}
    \mathbf{o} &= \text{SS2D}(\mathbf{i})
\end{aligned}
\end{equation}
where $\mathbf{i}$ denotes the 2D image input and $\mathbf{o}$ represents the scanned output. For more details of SS2D, interested readers are encouraged to the original work \cite{liu2024vmamba}.

\subsection{Network Architecture}
As depicted in Fig. \ref{fig:pipeline}, the proposed method comprises of Mamba-based backbone \cite{liu2024vmamba}, Cross-Modal Fusion Mamba (CMM), and Multi-level Refinement (MR) decoder. Specifically, for the input RGB and depth, we first adopt the Mamba-based backbone to extract modal-specific features, respectively. Then, we perform CMM to fuse the hierarchical RGB and depth features to obtain RGB-D features that have interacted between two modalities. Finally, we utilize the MR to aggregate the fused RGB-D features for the final prediction. The architecture is detailed below.

\textbf{Encoder.}
We feed the RGB and depth to the dual-stream Mamba-based backbone to capture multilevel features. The extracted RGB features are expressed as $\bm{F_i^{r}}$ while depth features are $\bm{F_i^{d}}$ ($i\in \{1,2,3,4,5\}$), as shown in Fig. \ref{fig:pipeline}. Different from previous methods that adopt CNN-based or Transformer-based backbone, we are the first to adopt Mamba-based backbone\cite{liu2024vmamba} into the field of RGB-D SOD.  

\textbf{Cross-Modal Fusion Mamba.}
Generally, the CMM takes the RGB features $\bm{F_i^r}$ and depth features $\bm{F_i^d}$ as inputs and outputs the fused RGB-D features $\bm{F_i}$. The proposed cross-modal fusion Mamba aims to model the inter-modal correlation and enhance the self-modality features. As illustrated in Fig. \ref{fig:CMM}, we first feed the extracted depth features and RGB features into the Mamba block for self-modal feature enhancement, respectively. Meanwhile, we concatenate the depth and RGB features to model their inter-modal correlation. This operation enhances cross-modal correlations by incorporating information from different modalities, thereby enriching the diversity of channel features. Then, the enhanced RGB and depth features are gated by inter-modal correlation features, encouraging complementary feature learning. In the end, we fused the processed RGB and depth features via element-wise addition, resulting in the image-like RGB-D features $\bm{F_i}$.

\begin{figure}[t]
	\centering
	\includegraphics[width=0.95\linewidth]{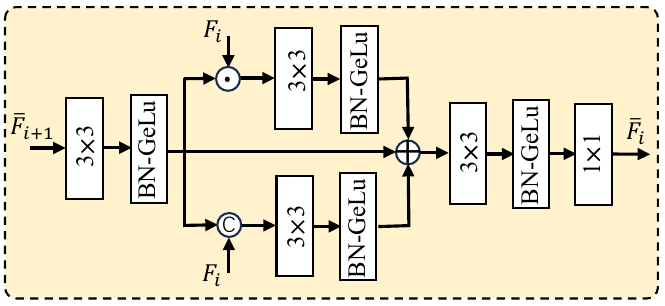}  \\
	\caption{Details of Multi-level Refinement module.}
	\label{fig:MRM}
\end{figure}

\textbf{Decoder.} We employ the Multi-level Refinement (MR) module to aggregate RGB-D features, and the processed features are denoted as $\bm{\bar{F_i}}$. As depicted in Fig. \ref{fig:MRM}, the purpose of the MR module is to effectively utilize the multi-level RGB-D features for salient object detection. We first upsampling the input feature $\bm{\bar{F}_{i+1}}$. The resulting features are separately utilized to enhance its next stage features by element-wise multiplication and concatenation, as the effectiveness is proven by previous works\cite{zeng2023compensated, zhong2024magnet}. Finally, we merge the enhanced features by element-wise addition, followed by convolution with kernel size $3 \times 3$ and $1\times1$. The aggregated features at each stage are denoted as $\bm{\bar{F}_{i}}$.

\subsection{Visual Mamba Block}
The role of the backbone network is to capture the hierarchical features from the RGB and depth inputs. Specifically, the inputs $\bm{RGB} \in \mathbb{R}^{3 \times H \times W}$ and $\bm{Depth} \in \mathbb{R}^{1 \times H \times W}$ are first divided into patches by a Patch Partition module so that we can obtain an initial feature representation with dimension of $C_1 \times \frac{H}{4} \times \frac{W}{4}$. Subsequently, we employ several similar operations to capture hierarchical features with dimension of $C_1 \times \frac{H}{4} \times \frac{W}{4}$, $C_2 \times \frac{H}{8} \times \frac{W}{8}$, $C_3 \times \frac{H}{16} \times \frac{W}{16}$, and $C_4 \times \frac{H}{32} \times \frac{W}{32}$, where $C_i$ denote the feature channel dimension. They are set to $96, 192, 384$, and $768$, respectively. Each operation consists of a down-sampling step (in addition to the first phase) followed by multiple Visual Mamba (VM) blocks as shown on the right side of Fig \ref{fig:pipeline}. 
\begin{figure}[t]
	\centering
	\includegraphics[width=1\linewidth]{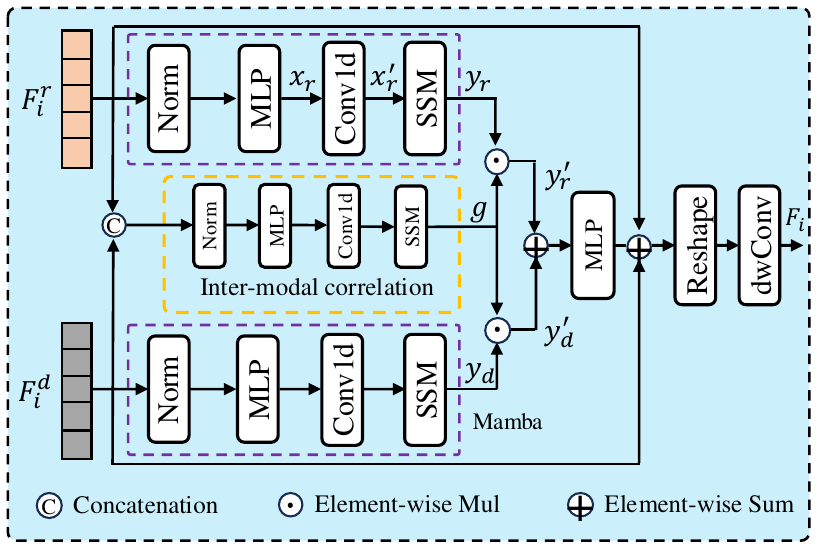}  \\
	\caption{Architecture of the proposed cross-modal fusion Mamba module.}
	\label{fig:CMM}
\end{figure}

The VM block is the visual counterpart for 2D image feature representation, while the initial Mamba block is used to process the 1D language sequence. More concretely, given an input sequence $\bm{z}$, the calculation of the entire VM block is divided into two parts, the first one is computed as:

\begin{equation}
\begin{aligned}
   \bm{\hat{z}} &= \text{SiLU}(\text{DWConv}(\text{Linear}(\text{LN}(\bm{z})))),
    \\
    \bm{\bar{z}} &= \text{Linear}(\text{LN}(\text{SS2D}(\bm{\hat{z}}))),
    \\
    \bm{z'} &= \bm{\bar{z} + z},
\end{aligned}
\label{eq6}
\end{equation}
where the 2D-Selective-Scan (SS2D) module introduces a four-dimensional scan for richer feature sources to achieve global receptive fields. The LN, DWConv, and SiLU denote LayerNorm, depth-wise separable convolution, and SiLU activation function, respectively. In the another step, the $\mathbf{z'}$ is further calculated as:
\begin{equation}
\begin{aligned}
    \bm{z''} &= \text{FNN}(\text{Linear}(\text{LN}(\bm{z'}))),
    \\
    \bm{y} &= \bm{z'' + z'},
\end{aligned}
\label{eq7}
\end{equation}
where FFN denotes feed forward network \cite{liu2024vmamba}.

\begin{table*}[t]
\scriptsize
        \renewcommand\arraystretch{1.4}   
        \caption
	{\textbf{Quantitative experiments on six challenging datasets.} The top three results are highlighted in \textcolor{red}{\textbf{red}}, \textcolor{blue}{\textbf{blue}} and \textcolor{green}{\textbf{green}}, respectively. $\uparrow (\downarrow)$ shows a higher (lower) value is preferable. PubYear denotes the year a reference was published. For example, `24' represents a reference published in 2024. Params and FLOPs denote the parameters and Floating Point Operations, respectively.}  
        \label{tab:table1}
	\centering
        \Huge
        \begin{threeparttable}
        \resizebox{\textwidth}{!}{
	\setlength{\tabcolsep}{0.8mm}{
	\begin{tabular}{c|r|cc ccc cc cccc ccccc cc}
	\toprule[2pt]
        \specialrule{0em}{0pt}{0pt}
        \multicolumn{2}{c|}{\makecell{Method \\Journal\\ PubYear}}  
        & \makecell{D3Net \\TNNLS \\ 20  \cite{fan2020rethinking}}
        & \makecell{BBSNet\\ECCV \\20 \cite{9405467}} 
        & \makecell{UCNet \\TPAMI \\ 21 \cite{zhang2021bts} } 
        & \makecell{MobileSal \\TPAMI \\ 21 \cite{wu2021mobilesal}}   
        & \makecell{DFMNet \\ACM MM \\ 21 \cite{zhang2021depth}}
        & \makecell{DCMF \\TIP \\ 22 \cite{wang2022learning}}
        & \makecell{SSLSOD \\AAAI \\ 22 \cite{zhao2022self}}
        & \makecell{CIRNet\\TIP \\22 \cite{9428263}} 
        & \makecell{C2DFNet\\TMM \\ 22 \cite{9813422}} 
        & \makecell{CAVER\\TIP \\ 23 \cite{10015667}} 
        & \makecell{MMRNet\\PR \\ 23 \cite{fang2023m2rnet}} 
        & \makecell{PICR-Net\\ACM MM \\ 23 \cite{cong2023point}}
        & \makecell{FCFNet \\TCSVT \\ 23 \cite{zhang2023feature}} 
        & \makecell{HIDANet\\TIP \\ 23 \cite{2023HiDAnet}}
        & \makecell{AirSOD\\TCSVT \\ 24 \cite{10184101}} 
        & \makecell{GTransNet \\Neurocom \\ 24 \cite{fang2024grouptransnet}} 
        & \textbf{\textbf{Ours}}  \\
    \hline
    \multirow{4}{*}{\rotatebox{90}{STERE}} 
    & $F_\beta^{max} \uparrow $  &0.891&0.903&0.899&0.906&0.893&0.872&0.882&\textcolor{blue}{0.913}&0.892&0.911&\textcolor{blue}{0.913}&0.910&0.906&0.894&0.900&0.895&\textbf{\textcolor{red}{0.920}}\\
    & $E_\xi^{max} \uparrow$     &0.938&0.942&0.944&0.940&0.941&0.930&0.929&0.928&0.927&\textcolor{green}{0.949}&0.929&\textcolor{blue}{0.951}&0.947&0.940&0.939&0.928&\textbf{\textcolor{red}{0.955}} \\
    &$S_\alpha \uparrow$         &0.899&0.908&0.903&0.903&0.898&0.903&0.887&\textcolor{blue}{0.916}&0.902&\textcolor{green}{0.914}&0.899&0.913&0.906&0.897&0.895&0.908&\textbf{\textcolor{red}{0.924}} \\
    & MAE $\downarrow$           &0.046&0.041&0.039&0.041&0.045&0.043&0.047&0.037&0.038&0.033&0.042&\textcolor{green}{0.033}&0.038&0.042&0.043&\textcolor{blue}{0.032}&\textbf{\textcolor{red}{0.031}} \\
    \cline{2-19}
    & Rank $\downarrow$          & 16 & 6 & 7 & 8 & 14 & 15 & 17 & 5 & 12 & 2 & 10 & 3 & 4 & 11 & 12 & 9 & \textbf{1} \\
    \hline

    \multirow{5}{*}{\rotatebox{90}{NLPR}} 
    & $F_\beta^{max} \uparrow $  &0.897&0.918&0.915&0.916&0.908&0.875&0.899&\textcolor{green}{0.924}&0.899&0.921&0.921&\textcolor{blue}{0.930}&0.911&\textcolor{green}{0.924}&0.923&0.908&\textbf{\textcolor{red}{0.934}} \\
    & $E_\xi^{max} \uparrow$     &0.953&0.961&0.956&0.961&0.957&0.940&0.949&0.955&0.958&0.961&0.941&\textcolor{blue}{0.971}&0.960&\textcolor{green}{0.964}&0.963&0.961&\textbf{\textcolor{red}{0.973}} \\
    &$S_\alpha \uparrow$         &0.912&0.930&0.920&0.920&0.923&0.922&0.915&\textcolor{green}{0.934}&0.928&0.929&0.918&\textcolor{blue}{0.936}&0.924&0.929&0.924&0.928&\textbf{\textcolor{red}{0.941}} \\
    & MAE $\downarrow$           &0.030&0.023&0.025&0.025&0.026&0.029&0.028&0.027&0.021&0.022&0.033&\textcolor{blue}{0.019}&0.024&0.021&0.023&\textcolor{blue}{0.019}&\textbf{\textcolor{red}{0.017}} \\
    \cline{2-19}
    & Rank $\downarrow$          & 17 & 5 & 12 & 11 & 13 & 16 & 15 & 8 & 9 & 4 & 14 & 2 & 10 & 3 & 5 & 7 & \textbf{1} \\
    \hline

    \multirow{5}{*}{\rotatebox{90}{NJU2K}} 
    & $F_\beta^{max} \uparrow $  &0.900&0.920&0.908&0.914&0.910&0.888&0.902&\textcolor{blue}{0.927}&.899&\textcolor{green}{0.923}&0.922&\textcolor{green}{0.923}&\textcolor{green}{0.923}&0.922&0.918&0.921&\textbf{\textcolor{red}{0.937}} \\
    & $E_\xi^{max} \uparrow$     &0.950&0.949&0.936&0.942&0.947&0.925&0.939&0.925&0.919&0.951&0.904&\textcolor{blue}{0.954}&\textcolor{green}{0.953}&\textcolor{green}{0.953}&0.944&0.926&\textbf{\textcolor{red}{0.963}} \\
    &$S_\alpha \uparrow$         &0.900&0.921&0.897&0.905&0.906&0.913&0.903&\textcolor{blue}{0.925}&0.908&0.920&0.910&\textcolor{green}{0.922}&0.918&0.920&0.908&\textcolor{green}{0.922}&\textbf{\textcolor{red}{0.934}} \\
    & MAE $\downarrow$           &0.041&0.035&0.043&0.041&0.042&0.043&0.044&0.035&0.038&\textcolor{green}{0.031}&0.049&0.032&0.034&\textcolor{green}{0.031}&0.039&\textcolor{blue}{0.028}&\textbf{\textcolor{red}{0.027}} \\
    \cline{2-19}
    & Rank $\downarrow$          & 12 & 8 & 17 & 11 & 10 & 15 & 16 & 6 & 14 & 3 & 13 & 2 & 5 & 4 & 9 & 7 & \textbf{1} \\
    \hline
  
    \multirow{5}{*}{\rotatebox{90}{DES}} 
    & $F_\beta^{max} \uparrow $  &0.885&0.927&0.920&-&0.922&0.901&-&0.878&0.896&-&-&\textcolor{red}{0.939}&\textcolor{red}{0.939}&0.918&0.934&\textcolor{blue}{0.938}&\textbf{0.935} \\
    & $E_\xi^{max} \uparrow$     &0.946&0.966&0.976&-&0.972&0.967&-&0.952&0.955&-&-&\textcolor{green}{0.970}&\textcolor{blue}{0.980}&0.954&0.962&\textcolor{red}{0.982}&\textbf{0.969} \\
    & $S_\alpha \uparrow$        &0.898&0.933&0.934&-&0.931&0.880&-&0.913&0.914&-&-&\textcolor{blue}{0.940}&\textcolor{green}{0.939}&0.919&0.923&\textcolor{red}{0.944}&\textbf{0.937} \\
    & MAE $\downarrow$           &0.031&0.021&0.019&-&0.021&0.023&-&0.025&0.021&-&-&\textcolor{blue}{0.017}&\textcolor{blue}{0.017}&0.021&0.022&\textcolor{red}{0.014}&\textbf{0.019} \\
    \cline{2-19}
    & Rank $\downarrow$          & 13 & 7 & 5 & - & 6 & 11 & - & 12 & 10 & - & - & 3 & 2 & 9 & 8 & 1 & \textbf{4} \\
    \hline
    
    \multirow{5}{*}{\rotatebox{90}{DUTLF}} 
    & $F_\beta^{max} \uparrow $  &0.785&-&0.836&0.912&0.747&0.928&0.914&0.937&0.934&\textcolor{blue}{0.942}&0.925&0.923&0.931&0.935&0.920&\textcolor{green}{0.939}&\textbf{\textcolor{red}{0.947}} \\
    & $E_\xi^{max} \uparrow$     &0.829&-&0.904&0.950&0.844&0.950&0.941&0.952&\textcolor{green}{0.958}&\textcolor{blue}{0.964}&0.935&0.953&0.956&\textcolor{green}{0.958}&0.946&\textcolor{green}{0.958}&\textbf{\textcolor{red}{0.967}} \\
    &$S_\alpha \uparrow$         &0.815&-&0.863&0.896&0.791&0.920&0.909&\textcolor{green}{0.932}&0.930&0.931&0.903&0.913&0.924&0.925&0.891&\textcolor{blue}{0.935}&\textbf{\textcolor{red}{0.942}} \\
    & MAE $\downarrow$           &0.048&-&0.064&0.041&0.092&0.035&0.038&0.048&\textcolor{blue}{0.025}&0.028&0.042&0.033&0.032&0.028&0.048&\textcolor{red}{0.024}&\textbf{\textcolor{red}{0.024}} \\
    \cline{2-19}
    & Rank $\downarrow$          & 15 & - & 14 & 12 & 16 & 9 & 10 & 7 & 4 & 3 & 11 & 8 & 6 & 5 & 13 & 2 & \textbf{1} \\    
    \hline
       
    \multirow{5}{*}{\rotatebox{90}{SIP}} 
    & $F_\beta^{max} \uparrow $  &0.861&0.883&0.879&0.898&0.887&-&0.894&0.895&0.867&\textcolor{blue}{0.906}&\textcolor{green}{0.902}&0.894&-&0.894&0.887&0.895&\textbf{\textcolor{red}{0.914}} \\
    & $E_\xi^{max} \uparrow$     &0.909&0.922&0.919&0.916&0.926&-&\textcolor{green}{0.928}&0.917&0.915&\textcolor{blue}{0.933}&0.921&0.926&-&0.925&0.903&0.926&\textbf{\textcolor{red}{0.940}} \\
    &$S_\alpha \uparrow$         &0.860&0.879&0.875&0.873&0.883&-&\textcolor{green}{0.890}&0.888&0.872&\textcolor{blue}{0.893}&0.882&0.881&-&0.884&0.858&0.887&\textbf{\textcolor{red}{0.904}} \\
    & MAE $\downarrow$           &0.063&0.055&0.051&0.053&0.051&-&0.047&0.052&0.054&\textcolor{green}{0.042}&0.049&0.050&-&0.050&0.060&\textcolor{blue}{0.041}&\textbf{\textcolor{red}{0.040}} \\
    \cline{2-19}
    & Rank $\downarrow$          & 15 & 12 & 11 & 10 & 9 & 0 & 4 & 8 & 13 & 2 & 5 & 7 & 0 & 6 & 14 & 3 & \textbf{1} \\
    \hline
    \multicolumn{2}{c|}{Params(M) $\downarrow$} &43.2&49.8&31.3&6.5&8.5&58.9&74.2&103.2&47.5&55.8&172.0&112&-&525&2.4&431.6&78.9\\
    \hline
    \multicolumn{2}{c|}{FLOPs(G) $\downarrow$} &-&31.4&-&1.6&-&271.1&144&42.6&11.1&44.4&-&27.1&-&-&0.9&-&16.9\\
    \bottomrule[2pt]
    \end{tabular}}}
    \end{threeparttable}
\end{table*}

\subsection{Cross-modal Fusion Mamba}
Both RGB and depth maps play a significant role in RGB-D SOD, although RGB features include rich semantic cues, depth features contain more distinguished object layout information. Proper fusing RGB and depth features is very important in the RGB-D SOD. Thus, we propose a novel cross-modal fusion Mamba module. As depicted in Fig. \ref{fig:CMM}, structurally, the CMM module can be divided into two parts: self-modal enhancement and inter-modal correlation modeling.

Specifically, as shown in Fig. \ref{fig:CMM}, the CMM module takes RGB features $\bm{F_i^r} \in \mathbb{R}^{B \times N \times C}$, depth features $\bm{F_i^d} \in \mathbb{R}^{B \times N \times C}$ as input (The input RGB and depth features are transformed into token before feeding into our CMM). The input token sequence $\bm{F^r_i}$ undergoes initial normalization via layer normalization. Next, the normalized sequence is projected into $\boldsymbol{x_r} \in \mathbb{R}^{B \times N \times C}$ using a Multi-layer Perception (MLP). Following this, a 1D convolution layer with SiLU activation function is applied to process $\boldsymbol{x_r}$ and yield $\boldsymbol{x_r}^{\prime}$. Subsequently, the $\boldsymbol{x_r}^{\prime}$ undergoes a state sequence model (SSM) for long-range dependencies modeling and yields $\boldsymbol{y_r}$. The self-modality enhancement process via Mamba can be defined as:
\begin{equation}
    \bm{y_r} = \text{SSM}(\text{Conv1}(\text{MLP}(\text{LN}(\bm{F^r_i})))),
    \label{Mamba_r}
\end{equation}
where LN and Conv1 denote the LayerNorm layer and 1D convolution layer, respectively.
Also, we can enhance the depth features by learning its long-range dependencies:
\begin{equation}
    \bm{y_d} = \text{SSM}(\text{Conv1}(\text{MLP}(\text{LN}(\bm{F^d_i})))).
    \label{Mamba_d}
\end{equation}

Meanwhile, to model inter-modal correlation, we project features from two modalities into a shared space, employing gating mechanisms to encourage complementary feature learning. Given RGB features $\bm{F_i^r} $ and depth features $\bm{F_i^d}$, we concatenate them together and followed by a basic Mamba block to model inter-modal correlation. This operation enhances cross-modal correlations by incorporating information from distinct channels, thereby enriching the diversity of channel features. The inter-modal correlation modeling can be formulated as:
\begin{equation}
    \bm{g} = \text{SSM}(\text{Conv1}(\text{MLP}(\text{LN}(\text{Cat}(\bm{F^r_i}, \bm{F^d_i}))))),
    \label{Mamba_inter-modal}
\end{equation}
where Cat represent feature concatenation operation.
Furthermore, the $\bm{y_r}$ and $\bm{y_d}$ are gated by $\bm{g}$ to encourage complementary feature learning. This process can be formulated as:
\begin{equation}
    \begin{aligned}
    \bm{y'_r} = \bm{y_r} \odot \bm{g},
    \\
    \bm{y'_d} = \bm{y_d} \odot \bm{g}.
\end{aligned}
\end{equation}

Finally, the fusion of these cross-modal features involves element-wise multiplication and element-wise addition operations, followed by MLP, reshaping, and depth-wise convolution operations to obtain 2D RGB-D features. This process can be expressed as:
\begin{equation}
    \bm{F_i} = \text{dwConv}(\text{Reshape}(\text{MLP}(\bm{y'_r+y'_d})+\bm{F_i^r+F_i^d})) .
\end{equation}

\begin{figure*}[t]
	\centering
         \captionsetup{
        justification=justified,
        singlelinecheck=false }
	\includegraphics[width=1\linewidth]{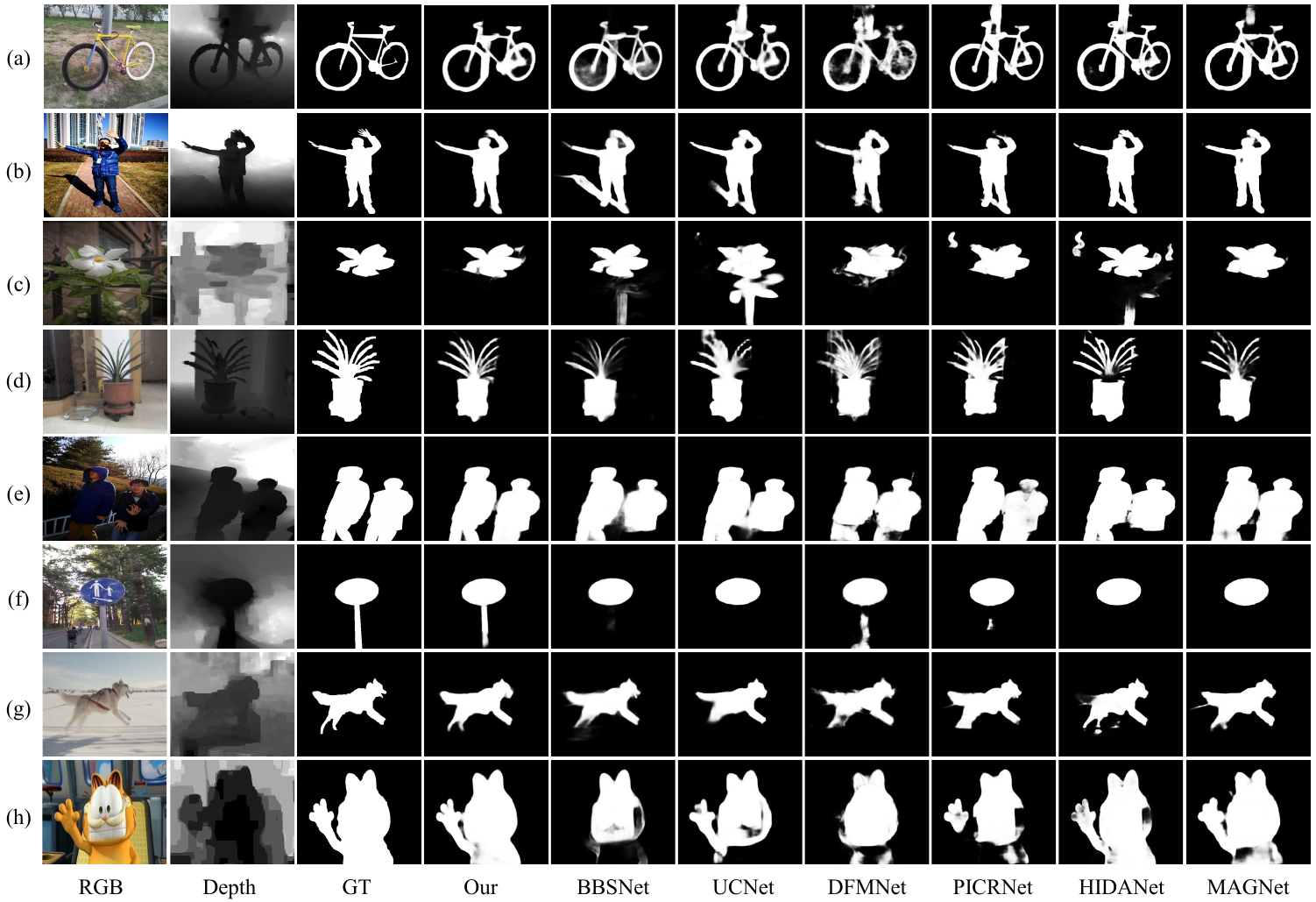}  \\
	\caption{Visual samples of different methods, including BBSNet \cite{9405467}, UCNet \cite{zhang2021bts}, DFMNet \cite{zhang2021depth}, PICRNet \cite{cong2023point}, HIDANet \cite{2023HiDAnet}, and MAGNet \cite{zhong2024magnet}. Our proposed method outperforms other SOTA models in various scenes, including common scenarios (line (a) and (b)), objects with sharp boundaries (line (c) and (d)),  multiple objects (line (e)), low contrast (line (f) and line (g)), and low-quality depth map((h)).}
	\label{fig:Figure7}
\end{figure*}
\section{EXPERIMENTS} \label{Section_4}
\subsection{Experimental details}
\textbf{Implementation setup.}
Our method is implemented on the PyTorch platform. For convenience, the depth and RGB inputs are both resized to $320\times320$. To augment the data, we conduct horizontal flipping and random cropping operations. We utilize a GeForce RTX 4090 GPU to train our model and test it on six different datasets. For training details, we set the epoch, batch size, and initial learning rate to 200, 10, and 0.0001, respectively. The Adam optimizer is adopted to train our model and the weight decay is set to 0.9 after epoch 60.

\textbf{Dataset.}
We perform experiments on six popular RGB-D SOD benchmark datasets: STERE\cite{niu2012leveraging}, NLPR\cite{peng2014rgbd}, NJU2K\cite{ju2014depth}, DES\cite{cheng2014depth}, SIP\cite{fan2020rethinking} and DUT\cite{piao2019depth}. NJU2K\cite{ju2014depth} is the largest RGB-D dataset, which includes 1985 RGB and depth images. NLPR\cite{peng2014rgbd} comprises 1000 RGB-D image pairs with the resolution of $640 \times 480$, and all of them are taken using a Microsoft Kinect. The STERE\cite{niu2012leveraging} dataset obtains 1000 stereoscopic RGB images and counterpart depth maps from the internet. DES\cite{cheng2014depth} is a smaller dataset that only contains 135 image pairs captured in some indoor scenes. SIP\cite{fan2020rethinking} also comprises 1000 RGB-D pairs but it was taken by a smartphone camera with a $992 \times 744$ resolution. In the DUT\cite{piao2019depth} dataset, including 1200 RGB and depth maps, there are several challenging scenarios, such as small objects, multiple objects, low contrast, and low-quality depth maps.

\begin{figure*}[t]
	\centering
	\includegraphics[width=1\textwidth]{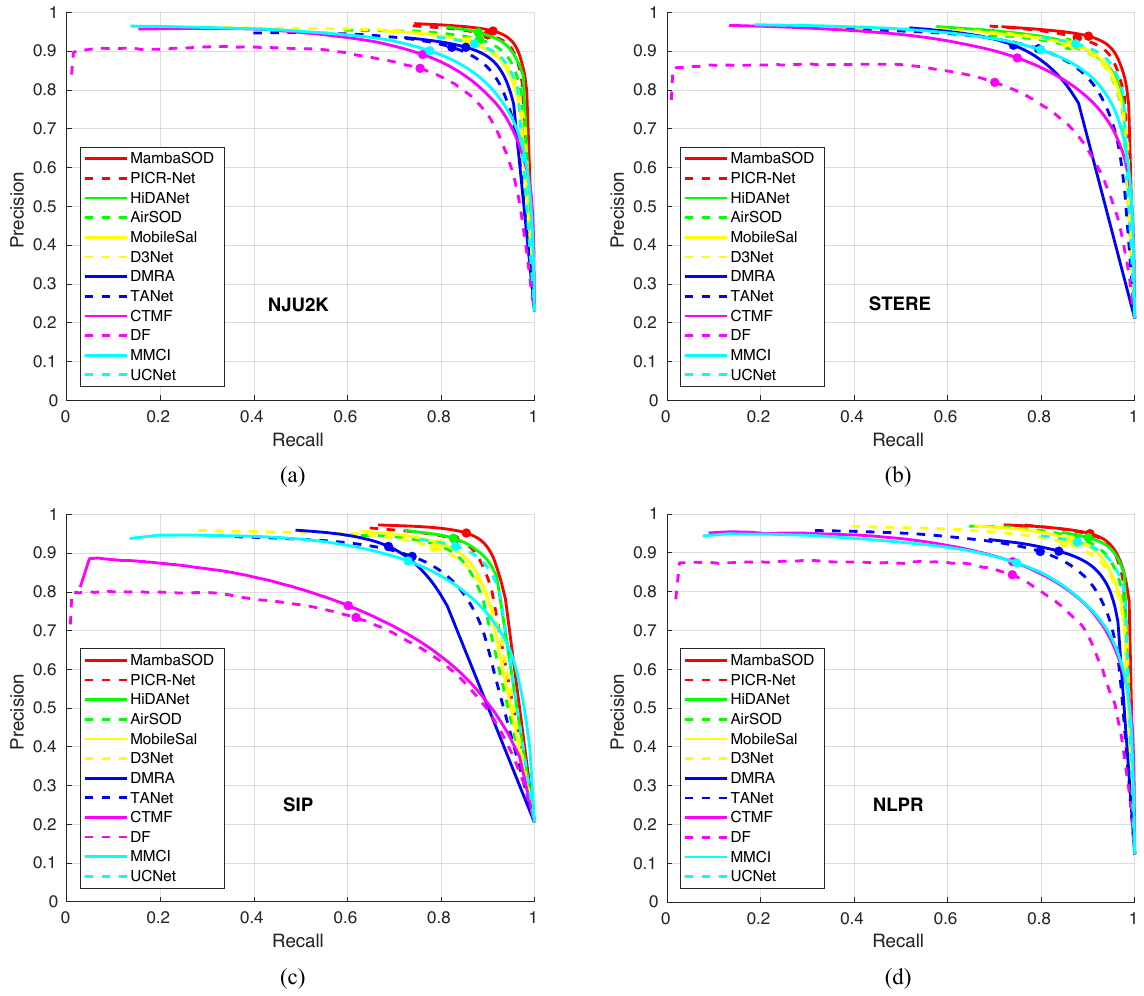}  \\
	\caption{Comparison of PR Curves for MambaSOD (solid red line) and other methods on four datasets:
 (a) PR curve on NJU2K dataset; (b) PR curve on STERE dataset; (c) PR curve on SIP dataset; (d) PR curve on NLPR dataset}  
 \label{fig:PR_Curve}
\end{figure*}

\textbf{Training and testing.}
Following previous studies \cite{9156838}, \cite{fan2020rethinking}, our method uses a specific subset for training data: 700 instances from NLPR and 1485 samples from NJU2K. The rest RGB-D pairs in NJU2K and NLPR, and all of the STERE, DES, and SIP datasets are treated as test data. Furthermore, for the DUTLF dataset, we train and test separately.

\textbf{Training Loss.}
During the training stage, we use the Binary Cross-Entropy (BCE) as the loss function, which measures the difference between the predicted probability distribution and the actual labels. It is defined as:
\begin{equation}
    \mathcal{L}_{\text{BCE}(P, G)} =  G \cdot \log(P) + (1 - G) \cdot \log(1 - P)
\end{equation}
where the P denotes the probability value output, and the G denotes the ground truth.

As our model adopts multi-level supervision, the overall loss can be shown below: 
\begin{equation}
    \mathcal{L}_{total} = \sum_{i=1} \mathcal{L}_{{BCE}(P_i,G)}
\end{equation}
where $P_i$ ($i\in \{1,2,3,4,5\}$) denotes the $i_{th}$ level predicted salient map in the decoder stage.

\textbf{Evaluation metrics.}
We quantitatively assess the performance of our method on four prevailing metrics: F-measure ($F_\beta$) \cite{achanta2009frequency}, E-measure ($E_\xi$) \cite{fan2018enhanced}, S-measure ($S_\alpha$) \cite{fan2017structure} and Mean Absolute Error (\text{MAE}) \cite{perazzi2012saliency}. Besides, we also calculate the Precision-Recall (PR) curves to show more comprehensive results.

The PR curve is used to assess a method's performance by plotting Precision against Recall at different threshold settings generally, if the curve is closer to the top-right area (i.e. higher precision and higher recall), which means a bigger area under the curve, the better model’s result. 

In the PR curve, Precision and Recall are the two main performance metrics. Precision denotes the proportion of positive cases that are really positive, which is computed as:

\begin{equation}
    \text{Precision} = \frac{\text{TP}}{\text{FP} + \text{TP}}
\end{equation}
where the True Positive (TP) is the correctly predicted positive case and the False Positive (FP) is the incorrectly predicted positive sample.

\begin{equation}
    \text{Recall} = \frac{\text{TP}}{\text{TP} + \text{FN}}
\end{equation}
where the False Negative (FN) is the incorrectly predicted negative case.

The F-measure ($F_\beta$) considers the weighted harmonic mean between Precision and Recall, which can be calculated below:
\begin{equation}
    F_{\beta} = \frac{(1 + \beta^2) \cdot \text{Precision} \cdot \text{Recall}}{\beta^2 \cdot \text{Precision} + \text{Recall}},
\end{equation}
where the default value of $\beta^2$ is 0.3 because of the significance of the precision, as advised by \cite{niu2012leveraging}.

The E-measure ($E_\xi$) describes a metric for computing the degree of similarity between the ground truth and the predicted salient map:
\begin{equation}
    E_{\xi} = \frac{1}{W \times H} \sum_{x=1}^{W} \sum_{y=1}^{H} f\left(\frac{2 \varphi_{GT} \circ \varphi_{FM}}{\varphi_{GT} \circ \varphi_{GT} + \varphi_{FM} \circ \varphi_{FM}}\right),
\end{equation}
where $W \times H$ represents the size of the image, and the $\varphi_{FM}$ and $\varphi_{GT}$ represent the feature representations of the ground truth and the predicted salient map, respectively.

The S-measure ($S_\alpha$) shows the structural similarity between the Ground Truth and the predicted saliency map, which is calculated as:
\begin{equation}
    S_{\alpha} = \alpha \ast S_{o} + (1 - \alpha) \ast S_{a},
\end{equation}
where the default value of $\alpha$ is 0.5 in order to balance the  area similarity $S_{a}$ and object similarity $S_{o}$, as advised by \cite{fan2017structure}.

The mean absolute error (\text{MAE}) is used to compute the error between the output of the model and the actual labels by summing and averaging all pixel errors over the entire image, which is calculated as:
\begin{equation}
    \text{MAE} = \frac{1}{W \times H} \sum_{i=1}^{W} \sum_{j=1}^{H} \left| P(i, j) - G(i, j) \right|,
\end{equation}
where $P(i, j)$ denotes the output value and $G(i, j)$ denotes the actual value.

\begin{table*}[t]
    \captionsetup{
        justification=justified,
        singlelinecheck=false 
    }
    \centering
    \renewcommand\arraystretch{1}
    \setlength{\tabcolsep}{2.2pt}
    \caption{Ablation experiments about the selection of feature extraction backbone. We adopt two widely-used feature extracted networks, ResNet50\cite{ResNet} and Swin Transformer\cite{liu2021swin} for comparison with our Mamba-based backbone. `t' denotes the tiny version.}
    \begin{tabular}{c|cccc|cccc|cccc|cccc} %
    \toprule[2pt]
    \multirow{2}{*}{\makecell{Backbone}} & \multicolumn{4}{c|}{NLPR} & \multicolumn{4}{c|}{STERE} & \multicolumn{4}{c|}{SIP} & \multicolumn{4}{c}{DES} \\
    \cline{2-17}
    & $F_\beta^{max} \uparrow$ & $E_\xi^{max} \uparrow$ & $S_\alpha \uparrow$ & MAE $\downarrow$
    & $F_\beta^{max} \uparrow$ & $E_\xi^{max} \uparrow$ & $S_\alpha \uparrow$ & MAE $\downarrow$
    & $F_\beta^{max} \uparrow$ & $E_\xi^{max} \uparrow$ & $S_\alpha \uparrow$ & MAE $\downarrow$ 
    & $F_\beta^{max} \uparrow$ & $E_\xi^{max} \uparrow$ & $S_\alpha \uparrow$ & MAE $\downarrow$\\
    \hline
    ResNet50 & 0.912 & 0.958 & 0.924 & 0.024 & 0.897 & 0.941 & 0.904 & 0.040 & 0.877 & 0.916 & 0.873 & 0.055 & 0.898 & 0.950 & 0.912 & 0.025 \\
    Swin-t & 0.915 & 0.964 & 0.929 & 0.023 & 0.898 & 0.946 & 0.903 & 0.043 & 0.877 & 0.921 & 0.970 & 0.058 & 0.921 & 0.965 & 0.930 & 0.022\\

    Ours & \textbf{0.934} & \textbf{0.973} & \textbf{0.941} & \textbf{0.017} & \textbf{0.920} & \textbf{0.955} & \textbf{0.924} & \textbf{0.031} & \textbf{0.914} & \textbf{0.940} & \textbf{0.904} & \textbf{0.040}& \textbf{0.935} & \textbf{0.969} & \textbf{0.937} & \textbf{0.019}\\

    \bottomrule[2pt]
    \end{tabular}
    \label{tab:backbone ablation}
\end{table*}

\subsection{Comparison with State-of-the-art methods}
\textbf{Contenders.}
We perform experiments on our and another sixteen SOTA methods to compare their performance. To be fair, the results we have shown are produced using authorized codes or supplied by the authors' publication papers if there is no code available.

\textbf{Quantitative Evaluation.} Table \ref{tab:table1} presents the quantitative results of our approach and other SOTA methods on sixteen popular datasets. Overall, compared to the SOTA methods, the numerous results illustrate that our approach achieves more satisfactory performance. Specifically, our performance ranks number one on the STERE\cite{niu2012leveraging}, NLPR\cite{peng2014rgbd}, NJU2K\cite{ju2014depth}, SIP\cite{fan2020rethinking} and DUTLF\cite{piao2019depth} datasets. Additionally, our MambaSOD achieves the fourth-best performance on DES\cite{cheng2014depth} dataset. The reason for this result is that the DES has a small number of samples with partially duplicated salient targets, which leads to performance fluctuations in our approach. Nevertheless, our method outperforms the majority of other SOTA methods and is only a few thousandths of a degree away from the top three best methods. The excellent results above are primarily due to our method's ability to extract local and also global features effectively and conduct valid self-enhancement and cross-modal interaction to them.

\textbf{Qualitative Evaluation.} Fig. \ref{fig:Figure7} presents a qualitative analysis of our method against other methods in several common and challenging scenarios. In sum, the results show our proposed approach segments the salient objects effectively, owing to its capability to effectively extract and fuse RGB and depth features. First of all, our method achieves better performance in some simple scenarios, like lines (a) and (b). Then, in the case of objects with sharp boundaries (lines (c) and (d)), our method outperforms in providing more details to identify the salient object with full edges. Next, for the images containing multiple objects, shown as line (e), our approach can identify salient objects effectively as our feature extractor can obtain richer cues related to RGB and depth maps. In complex environments with low contrast (lines (f) and (g)), our method overcomes the challenge and accurately segments objects thanks to the enhanced RGB-D spatial features. Lastly, the example of line (h) demonstrates our method's capability of distinguishing the salient object with low-quality depth maps, owing to the cross-modal interaction between RGB and depth. 

\textbf{PR Curve.} Furthermore, we draw different methods' PR curves on four datasets, as shown in Fig. \ref{fig:PR_Curve}, which includes our method (MambaSOD) and sixteen SOTA methods. According to the results of the PR curves, the performance of most CNN-based methods is worse than ours. This is mainly attributed their inability to model long-range dependencies, leading to performance degradation. Some latest methods (Published in 2024 and 2023), such as PICR-Net\cite{cong2023point}, AirSOD\cite{10184101}, HiDNet\cite{2023HiDAnet} achieved similar Precision and Recall with ours, but our MambaSOD still a slightly better than these methods, it can be owed to the effective inter-modal correlation modeling of our CMM module. Overall, the PR curves on four datasets prove the validity and superiority of our method.
Both the quantitative and qualitative experiments prove the validity of our proposed MambaSOD.

\textbf{Parameters and FLOPs Analysis.}
We also compare multiple SOTA methods with ours in terms of parameters and FLOPs to evaluate the size and computational complexity of different models. As depicted in the bottom of Table \ref{tab:table1}, our model contains a medium level of parameters 78.9M and low FLOPs 16.9G, while achieving promising model performance. 

\subsection{Ablation Study} 
Several ablation experiments are conducted to assess our proposed method on NJU2K\cite{ju2014depth}, NLPR\cite{peng2014rgbd}, STERE\cite{niu2012leveraging}, DES\cite{cheng2014depth}, SIP\cite{fan2020rethinking} and DUT\cite{piao2019depth} datasets. 

\textbf{The effectiveness of our backbone network.}
To evaluate the effectiveness of different backbones, we utilize different backbone network in the encoder for comparison. Specifically,
we choose ResNet50\cite{ResNet}, Swin Transformer (tiny version)\cite{liu2021swin}, and our Mamba-based backbone as feature extractors in encoding stage, respectively. Then we compare the detection performance of these models by conducting ablation experiments on four common-used datasets, including NLPR\cite{peng2014rgbd}, STERE\cite{niu2012leveraging}, SIP\cite{fan2020rethinking}, and DES\cite{cheng2014depth}.

The experimental results in Table \ref{tab:backbone ablation} show that our Mamba-based backbone outperforms the other backbones in all datasets. Specifically, the $F_\beta^{max}$ metric of our feature extractor reaches 0.934, which is higher than ResNet-50's 0.912 and Swin-t's 0.915. In the dataset, STERE\cite{niu2012leveraging}, SIP\cite{fan2020rethinking}, both ResNet-50 and Swin-t fail to achieve accurate enough results, while our model still maintains better performance. This shows that the Mamba-based backbone is more robust in processing RGB and depth with different qualities. The outstanding performance of the backbone based on Mamba across multiple datasets can be attributed to its effective extraction of both local and global features, which is particularly important for RGB-D SOD.

\begin{table}[t]
    \captionsetup{
        justification=justified,
        singlelinecheck=false 
    }
	\caption{Ablation study about cross-modal fusion Mamba on DUT\cite{piao2019depth} dataset. `Model\_A' denotes a baseline model, in which a simple element-wise addition is used. `Model\_B' applies our CMM but removes the inter-modal branch. `Ours' shows a complete implementation of our CMM.}
	\centering
    \setlength{\tabcolsep}{0.8mm}{
	\begin{tabular}{ c c c c c c c c}
		\toprule[2pt]
		Module & Baseline & Variant & CMM & $F_\beta^{max} \uparrow$ &  $E_\xi^{max} \downarrow$ & $S_\alpha \uparrow$ & $MAE \downarrow$\\
		\midrule[2pt]
		Model\_A & \checkmark & -  &  - & 0.939 & 0.961 & 0.932 & 0.030\\
		Model\_B & - & \checkmark & - & 0.943 & 0.964 & 0.938 & 0.026 \\
		\textbf{Ours} & - & - & \checkmark  & \textbf{0.947} & \textbf{0.967} & \textbf{0.942} &\textbf{0.024}\\
		\bottomrule[2pt]
	\end{tabular}
    }
    \label{tab:CMM ablation}
\end{table}

\textbf{The effectiveness of our cross-modal fusion.}
We further perform experiments to demonstrate the validity of our cross-modal fusion module. First, we compare three strategies of feature fusion, including Model\_A (conducting simple element-wise addition), Model\_B (applying CMM but removing the inter-modal branch), and the complete CMM module. As shown in table \ref{tab:CMM ablation}, Model\_A first provides the base performance, which shows an initial benchmark for us to evaluate the benefits of the CMM. Compared with Model\_A, the performance improvement of Model\_B  demonstrates that even in the absence of the inter-modal correlation branch, the CMM can still provide a performance boost. Our method gets the best performance, which proves the inter-modal branch in further enhancing the validity of our cross-modal fusion module. Specifically, in Model\_A, Model\_B, and our model, the $F_\beta^{max}$ increases from 0.939, 0.943 to 0.947, respectively. 

\begin{figure}[t]
	\centering
        \captionsetup{
        justification=justified,
        singlelinecheck=false 
      }
	\includegraphics[width=1\linewidth]{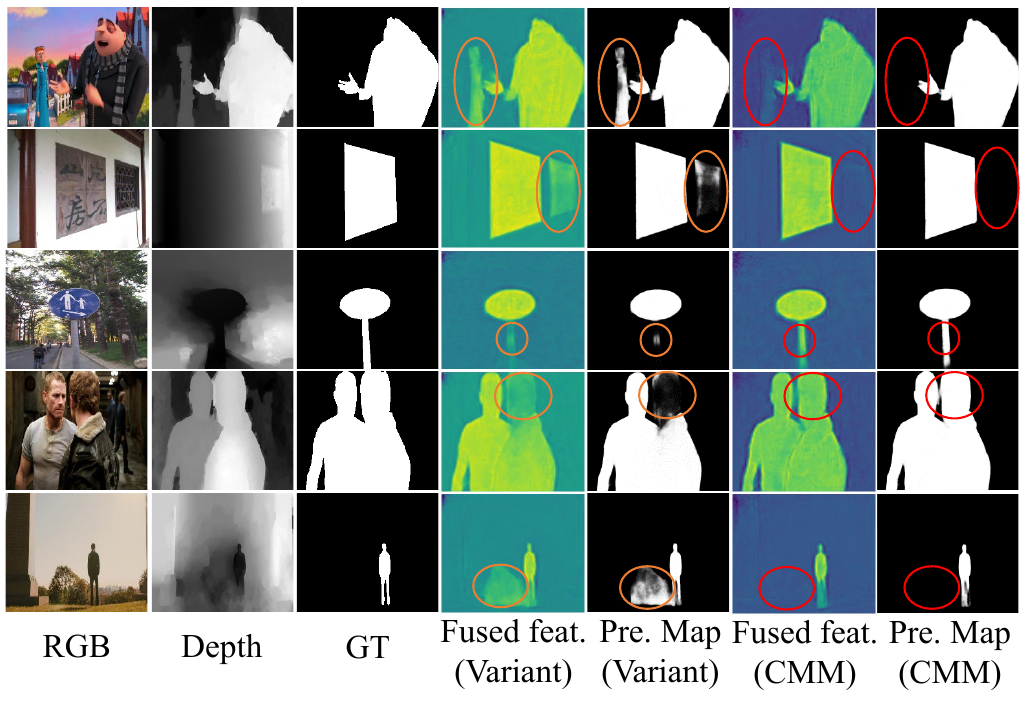}  \\
	\caption{Some visual comparison with and without the inter-modal correlation branch in our CMM module. `Fused feat.' denotes the processed RGB-D features undergoing fusion operation and `Pre. Map' shows the model's final prediction of the salient object. `Variant' represents our CMM module without the inter-modal branch.}
	\label{fig:CMM_ablation}
\end{figure}

Additionally, we conduct a qualitative visualization analysis of the CMM module with and without (variant) the inter-modal branch. As shown in Fig. \ref{fig:CMM_ablation}, we can conclude that the inter-modal interaction between different modalities is particularly crucial. Specifically, it can overcome the limitation brought by single-modal or simple multi-modal addition operations. More importantly, inter-modal correlation modeling enables the model to capture more useful information and discard useless or even intrusive cues. As the first line of Fig. \ref{fig:CMM_ablation} an example, the depth map contains some disturbing information so that the variant of CMM identifies the women as salient targets as well. Thanks to the cross-modal interaction, the complete CMM successfully suppressed the misleading information by utilizing complementary cues from the RGB.

\textbf{Comparison experiments on feature fusion module.} To further demonstrate the superiority of our CMM, we replace our feature fusion module with ones from other SOTA methods, including MobileSal (TPAMI'21) \cite{wu2021mobilesal}, DCMF\cite{wang2022learning} (TIP'22), HIDANet (TIP'23) \cite{2023HiDAnet} and AirSOD (TCSVT'24)\cite{10184101}. As shown in table \ref{tab:fusion method}, our crosss-modal fusion Mamba modules perform better than the ones used in other approaches. Overall, our network achieves higher model performance while also owning medium parameters and FLOPs.
\begin{table}[t]
    \captionsetup{
        justification=justified,
        singlelinecheck=false 
    }
	\caption{Comparison experiments among fusion modules of different methods on DUT\cite{piao2019depth} dataset. Specifically, we replace the CMM with the fusion module from SOTA methods.}
	\centering
    \setlength{\tabcolsep}{0.8mm}{
	\begin{tabular}{ c c c c c c c}
		\toprule[2pt]
		Method & Params(M) $\downarrow$ & FLOPs(G) $\downarrow$ &$F_\beta^{max} \uparrow$ &  $E_\xi^{max} \downarrow$ & $S_\alpha \uparrow$ & $MAE \downarrow$\\
		\midrule[2pt]
		MobileSal & 84.3 & 16.8 & 0.944 & 0.964 & 0.938 & 0.026\\
		DCMF & 78.5 & 16.9 & 0.946 & 0.966 & 0.940 & 0.025 \\
            HIDANet & 83.0 & 19.0 & 0.944 & 0.964 & 0.939 & 0.025 \\
            AirSOD & 75.8 & 15.1 & 0.940 & 0.961 & 0.933 & 0.028 \\
		\textbf{Ours} & \textbf{78.9} & \textbf{16.9} & \textbf{0.947} & \textbf{0.967} & \textbf{0.942} &\textbf{0.024}\\
		\bottomrule[2pt]
	\end{tabular}
    }
    \label{tab:fusion method}
\end{table}

\begin{figure}[t]
	\centering
	\includegraphics[width=1\linewidth]{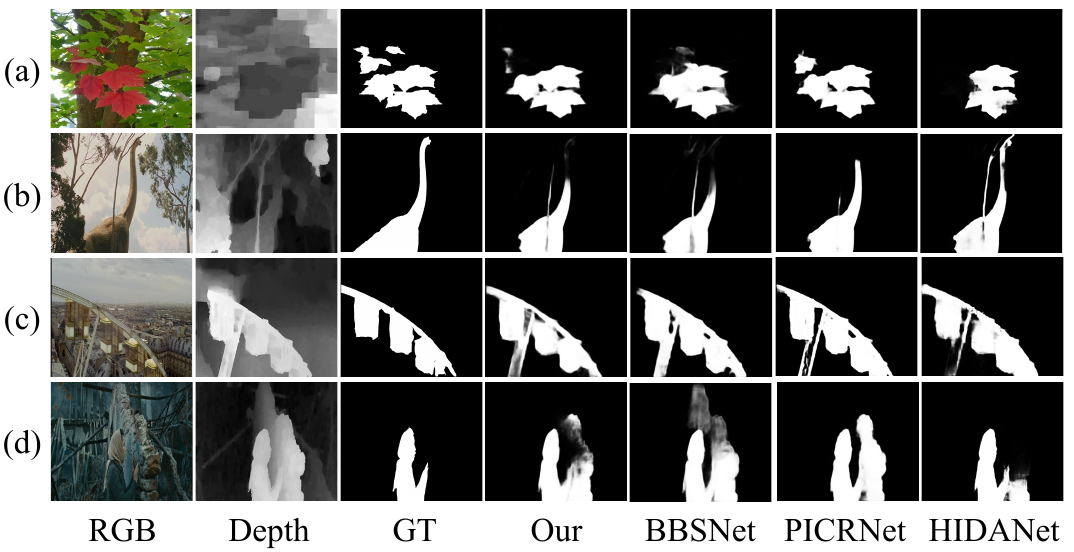}  \\
	\caption{Failure cased: some cases in which all SOTA and our methods fail to perform well.}
	\label{fig:failture_case}
\end{figure}

\textbf{Failure Cases.}
We show some representative failure examples to analyze the inadequacy of our method. As depicted in Fig. \ref{fig:failture_case}, some low-quality depth maps in rows (a) and (b), lead to the model can not accurately distinguish the salient objects from the background; While the RGB image information in the (c) and (d) rows is too redundant and confusing, which also increases the difficulty of the method in recognizing salient targets. In these scenes, our proposed method fails to accurately segment the salient object to some extent. It's worth noting that even some SOTA methods, like BBSNet \cite{9405467}, PICR-Net \cite{cong2023point}, and HIDANet \cite{2023HiDAnet} also failed to achieve satisfactory performance.

\section{CONCLUSION} \label{Section_5}
We propose a novel method, named MambaSOD, for the prevailing RGB-D SOD. Firstly, different from previous works that use ResNet or Transformer as the backbone, we are the first to introduce a Mamba-based backbone into the RGB-D SOD field, which is beneficial from long sequence modeling capability. Secondly, drawing inspiration from the State Space model, we propose a cross-modal fusion Mamba to effectively merge the features of RGB and depth, which contributes to the overall improvement in model performance. The proposed method achieves global feature extraction and facilitates cross-modal information exchange with linear complexity.
Our method is evaluated on six widely-used datasets and the experimental results demonstrate that our MambaSOD can achieve superior performance against other SOTA methods. For example, our method achieves the best overall performance in STERE, NLPR, NJU2K, DUTLF, and SIP datasets. The experimental results verify the effectiveness and superiority of our method. 
In the future, we aim to develop a more lightweight model while maintaining the performance of our current method.

\bibliographystyle{IEEEtran}
\normalem
\bibliography{REFERENCES}

\begin{IEEEbiography}[{\includegraphics[width=1in,height=1.25in,clip,keepaspectratio]{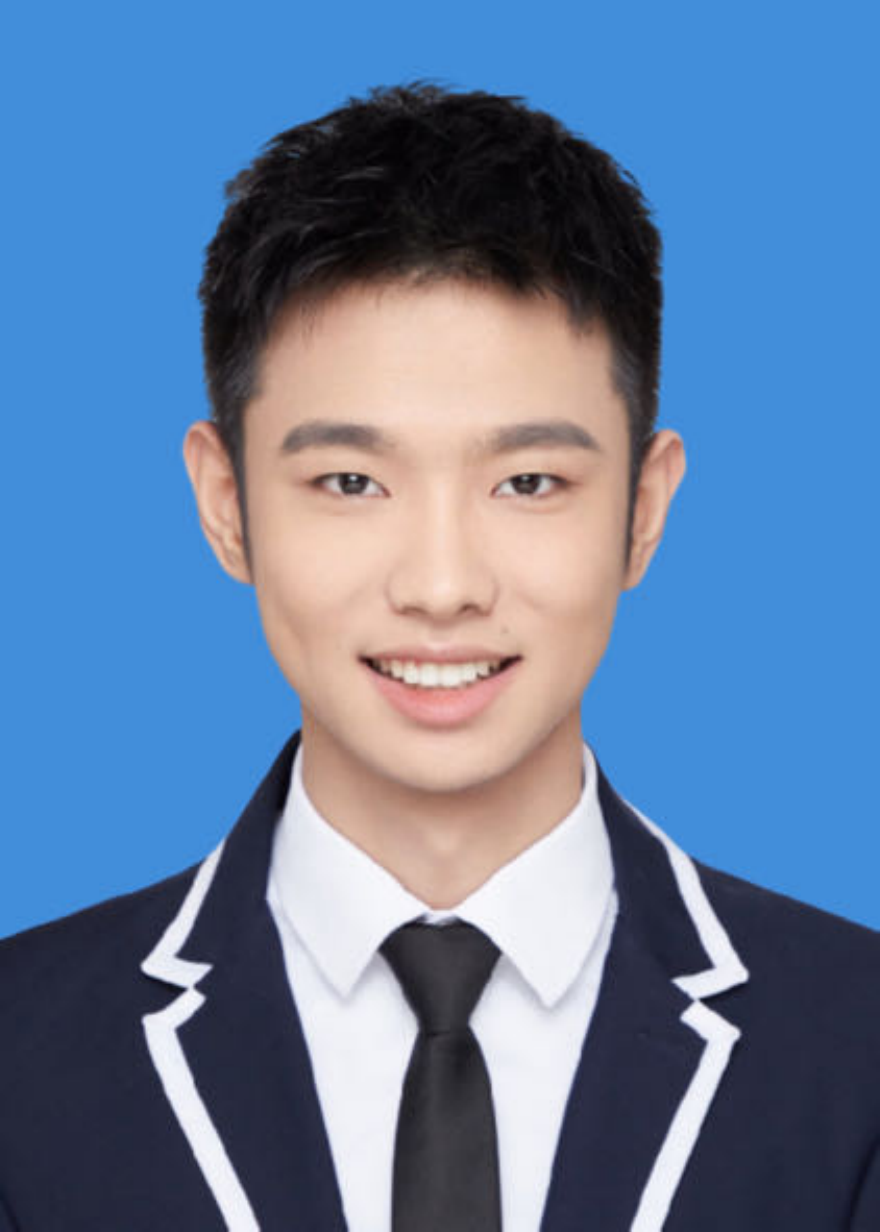}}]{Yue Zhan} is pursuing his M.Eng degree at the University of Hong Kong, Hong Kong SAR, China. His current research interests include computer vision, salient object detection, and large language model.
\end{IEEEbiography}

\begin{IEEEbiography}[{\includegraphics[width=1in,height=1.25in,clip,keepaspectratio]{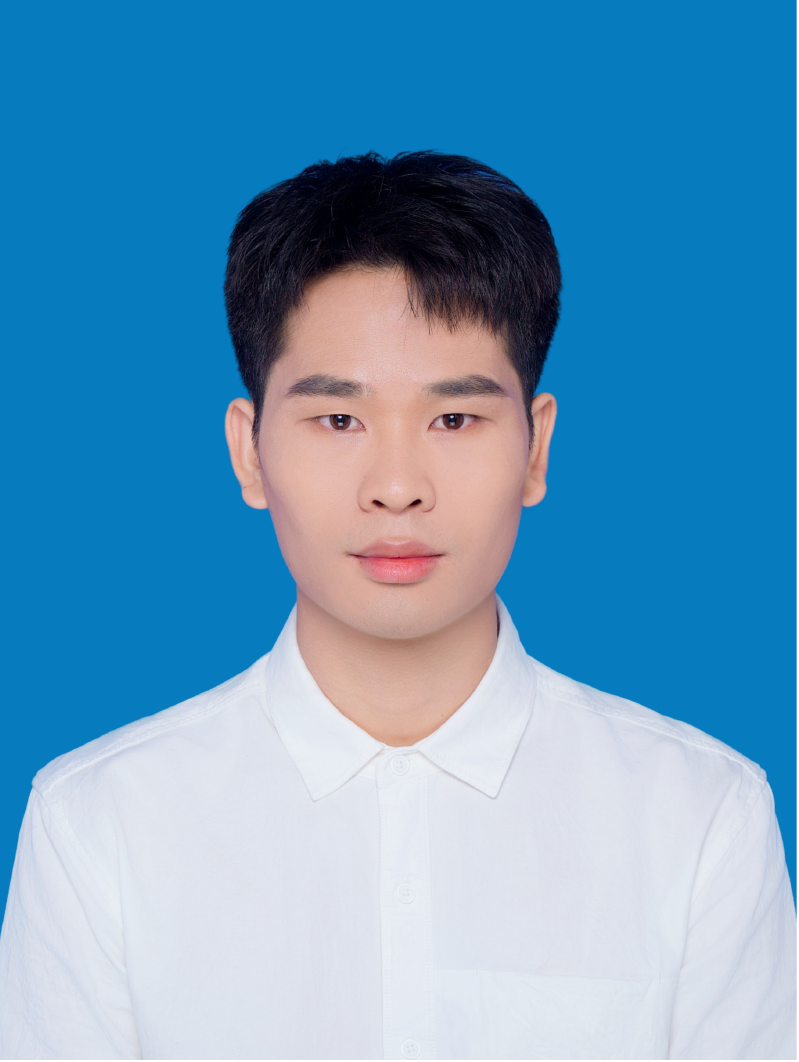}}]{Zhihong Zeng} is pursuing his PhD degree at Chongqing University, Chongqing, China. His current research interests include computer vision, deep learning and salient object detection.
\end{IEEEbiography}
\newpage
\begin{IEEEbiography}[{\includegraphics[width=1in,height=1.25in,clip,keepaspectratio]{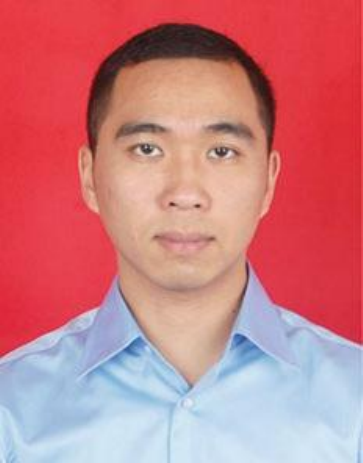}}]{Haijun Liu}
 is currently an associate professor in the School of Microelectronics and Communication Engineering at Chongqing University. He received the B.Eng, M.Eng and Ph.D degree from University of Electronic Science and Technology of China in 2011, 2014 and 2019, and has been a visiting scholar of Kyoto University from 2018 to 2019. His main research interests include manifold learning, metric learning, deep learning, subspace clustering and sparse representation in computer vision and machine learning, with focuses on human action detection and recognition, face detection and recognition, person detection and re-identification, and remote sensing image processing.
 He has served as a reviewer for many journals, including IEEE TIP, TMM, TCSVT, SPL, etc., and also the Editorial Board member of Journal of Artificial Intelligence and Systems.
\end{IEEEbiography}

\begin{IEEEbiography}[{\includegraphics[width=1in,height=1.25in,clip,keepaspectratio]{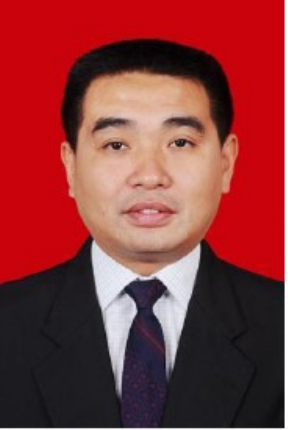}}]{Xiaoheng Tan} 
received the B.E. and Ph.D. degrees in electrical engineering from Chongqing University, Chongqing, China, in 1998 and 2003, respectively. He was a Visiting Scholar with the University of Queensland, Brisbane, Qld., Australia, from 2008 to 2009. He is currently a professor with the School of Microelectronics and Communication Engineering, Chongqing University. His current research interests include modern communications technologies and systems, communications signal processing, pattern recognition, and machine learning.
\end{IEEEbiography}

\begin{IEEEbiography}[{\includegraphics[width=1in,height=1.25in,clip,keepaspectratio]{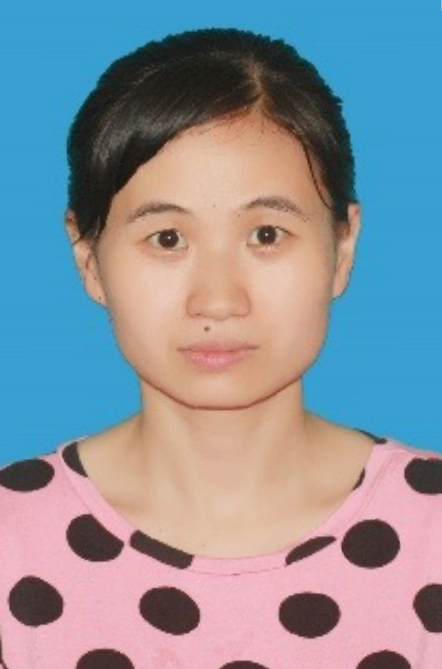}}]{Yinli Tian} 
received the Ph.D. degree in circuits and systems from Chongqing University, Chongqing, China, in 2022.
She is currently a Lecturer with the School of Software Engineering, Chongqing University of Posts and Telecommunications, Chongqing.
Her research interests include artificial intelligence, signal detection and analysis, and image analysis.
\end{IEEEbiography}

\end{document}